\documentclass[preprint, 12pt]{elsarticle}

\PassOptionsToPackage{pdfencoding=auto,pdfversion=1.7}{hyperref}
\usepackage[pdftex,hidelinks]{hyperref}

\usepackage{amssymb}
\usepackage{amsmath}
\usepackage{booktabs}
\usepackage{xcolor}

\usepackage{graphicx}
\usepackage{placeins}

\usepackage{multirow}
\journal{Engineering Applications of Artificial Intelligence}

\begin{document}

\begin{frontmatter}



\title{Fine-tuning of lightweight large language models for sentiment classification on heterogeneous financial textual data}


\author{Alvaro Paredes Amorin}
\affiliation{organization={International Business School, Zhejiang University},
            addressline={718 East Haizhou Rd.}, 
            city={Haining},
            postcode={314400}, 
            state={Zhejiang},
            country={China}}
\author{Andre Python}
\affiliation{organization={Center for Data Science, Zhejiang University},
            addressline={866 Yuhangtang Rd.}, 
            city={Hangzhou},
            postcode={310058}, 
            state={Zhejiang},
            country={China}}
\author{Christoph Weisser} 
\affiliation{organization={Centre for Statistics, Georg-August-Universität Göttingen},
            addressline={Humboldtallee 3}, 
            city={Goettingen},
            postcode={37073}, 
            state={Lower Saxony},
            country={Germany}}

\begin{abstract}
Large language models (LLMs) play an increasingly important role in financial markets analysis by capturing signals from complex and heterogeneous textual data sources, such as tweets, news articles, reports, and microblogs. 
However, their performance is dependent on large computational resources and proprietary datasets, which are costly, restricted, and therefore inaccessible to many researchers and practitioners. To reflect realistic situations we investigate the ability of lightweight open-source LLMs - smaller and publicly available models designed to operate with limited computational resources - to generalize sentiment understanding from financial datasets of varying sizes, sources, formats, and languages. We compare the benchmark finance natural language processing (NLP) model, FinBERT, and three open-source lightweight LLMs, DeepSeek-LLM 7B, Llama3 8B Instruct, and Qwen3 8B on five publicly available datasets: FinancialPhraseBank, Financial Question Answering, Gold News Sentiment, Twitter Sentiment and Chinese Finance Sentiment. We find that LLMs, specially Qwen3 8B and Llama3 8B, perform best in most scenarios, even from using only 5\% of the available training data. These results hold in zero-shot and few-shot learning scenarios. Our findings indicate that lightweight, open-source large language models (LLMs) constitute a cost-effective option, as they can achieve competitive performance on heterogeneous textual data even when trained on only a limited subset of the extensive annotated corpora that are typically deemed necessary.\\

\end{abstract}




\end{frontmatter}



\section{Introduction}
\label{Intro}


By capturing the prevailing attitude and mood of investors in the market, sentiment data play a critical role in financial market prediction, risk management, and financial modeling. In contrast to common financial market indicators that are derived from numerical data, sentiment is obtained from textual sources such as news headlines, analyst reports, and social networks to capture emotions of investors that can be associated with prices and volatility \cite{b18, b19}.

Natural language processing (NLP) models have been commonly applied to analyze emotions in various contexts and languages \cite{b5, b20, b24, b25, b26, b27}. Large language models (LLMs) have systematically outperformed NLP models in sentiment analysis tasks by enhancing contextual comprehension through the use of deep learning techniques \cite{b4, b5, b7, b8, b9}. However, the training process of LLMs is computationally expensive, as it requires very large datasets and computational power, and specialized processors such as graphics processing units (GPUs) for massive parallel processing, which are often costly. Furthermore, high-performing LLMs, such as OpenAI GPT4 or Google Gemini, are limited to inference via APIs. Although multilingual LLMs such as XLM-R and mBERT support cross-lingual generalization, they usually require extensive fine-tuning to perform well in low-resource or domain-specific tasks such as financial sentiment analysis. To account for real-life scenarios where data and computational resources are limited, we consider light-weight versions of LLMs (between 7 and 8 billion parameters), as they offer a favorable balance between capability and resource demands, achieving high performances and enabling fine-tuning on devices with limited GPU power.



We compared FinBERT, DeepSeek-LLM 7B, Llama3 8B and Qwen3 8B in their ability to generalize understanding of sentiment in financial domains from different sources, including financial news, social media content, and forum discussion. 
Since FinBERT is a transformer-based model previously fine-tuned on the FinancialPhraseBank and Financial Question Answering (FiQA) datasets, we include those datasets for comparison, alongside two distinct labeled sentiment datasets: the Gold Sentiment dataset, which exclusively contains financial news about gold, and the Twitter Financial News Sentiment dataset, which is a dataset containing an English-language annotated corpus of finance-related tweets. We extend these commonly used financial datasets with a sentiment data set in Chinese-language. This allows us to assess the ability of models to operate in so-called analytic languages such as Chinese, where word order and particles are used to express grammatical relationships. This contrasts with inflectional languages, such as English, where words change form to convey grammar (e.g., tense, number, or case). We applied a Role-Playing (RP) format to design prompts used to train and evaluate the LLMs. This approach, introduced by \cite{b16} appears more effective than vanilla prompts in various research works that applied LLMs in sentiment analysis \cite{b1, b17}.

\begin{figure}[t]
       \centering
          \includegraphics[width=1\textwidth]{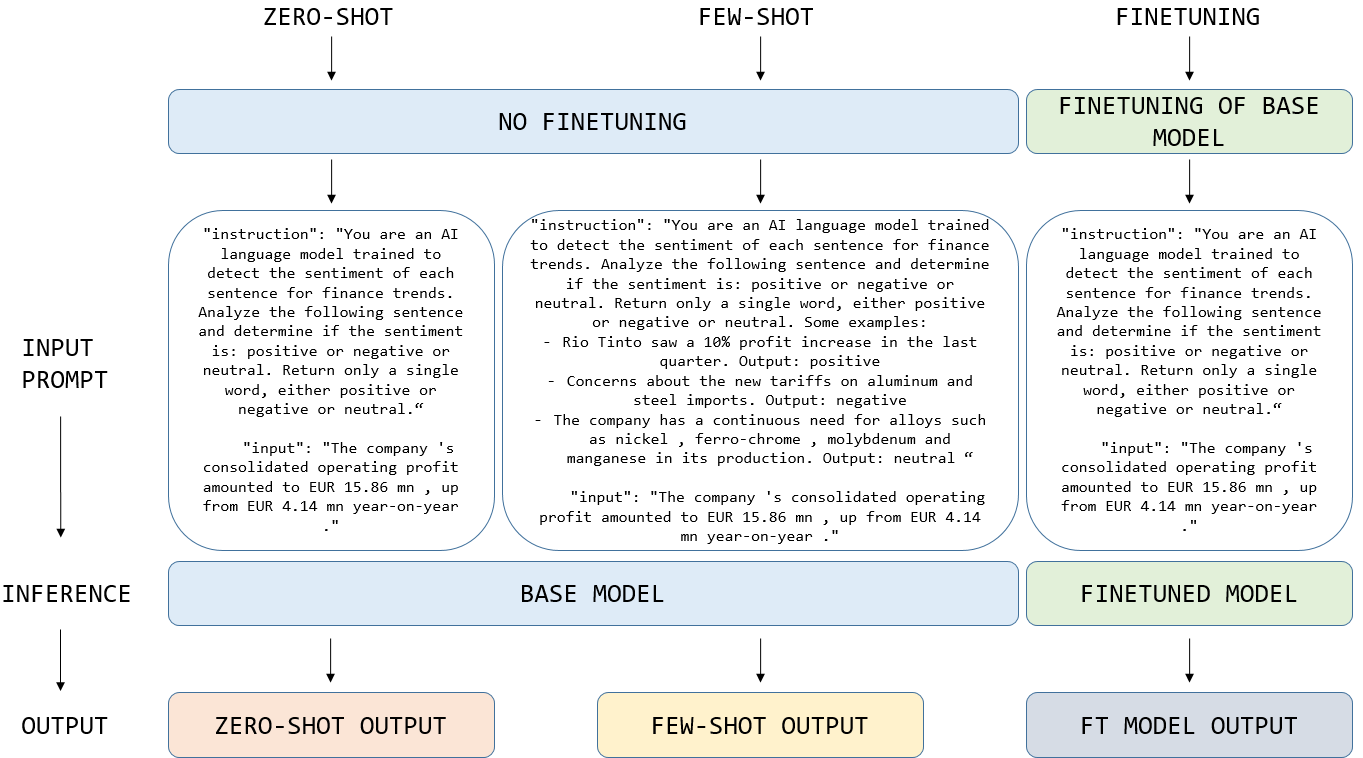}
          \caption{Differences between zero-shot, few-shot and fine-tuning. Zero-shot is using a prompt providing no examples while few-shot does, e.g. 3-shot means using prompts containing 3 examples. Fine-tuning consists of training a pre-trained model or base model with specific-task data.}
          \label{zeroshot_fewshot_figure}
\end{figure}

LLMs can improve their performance by further training on a specific dataset or task (supervised fine-tuning, SFT) or from additional prompt examples. In real-case scenarios, often only very limited or unspecialized training data may be available. Common methods to mitigate the problem of sparse training data include zero shot learning (ZSL) and few shot learning (FSL) \cite{b2, b5} (Figure \ref{zeroshot_fewshot_figure}). FSL consists of adding some examples inside the prompt of how the model should behave. For example, a 3-shot case refers to a prompt that includes 3 examples of input-output pairs. Both FSL and ZSL are computationally efficient as they do not require extensive datasets or resources for fine-tuning. 

To achieve effective training with varied datasets, we designed a pipeline that integrates financial sentiment datasets from different sources and media support, such as tweets, news headlines, or microblogs. We fine-tuned the models using the training splits of all available labeled datasets: FinancialPhraseBank, FiQA, Twitter Sentiment, and Gold Sentiment. The evaluation is then conducted on the respective test splits of each dataset. 


\section{Related Work}

\subsection{Sentiment Analysis with LLMs}

Recent developments in large language models (LLMs) have exerted a substantial influence on methodologies for sentiment analysis applied to financial textual data. Kumar et al. \cite{b2} fine-tuned LLaMA3 using only the FinancialPhraseBank (FPB) training dataset and achieved better predictive performance than with BERT-based models. Zhang et al. \cite{b7} adopted an instruction tuning method to fine-tune LLaMA 7B in various financial sentiment datasets. Zhang et al. \cite{b4} fine-tuned an LLaMA 7B model and implemented a retrieval augmented generation (RAG) module. Fatemi et al. \cite{b9} tested Flan-T5 Base, Large and XL models as well as ChatGPT turbo 3.5 in zero and few-shot learning settings. They further fine-tuned the Flan-T5 models using some of the previously introduced financial sentiment datasets, achieving better results than in zero-shot and few-shot learning scenarios. They also found that in zero and few-shot learning contexts, fine-tuned LLaMA3 8B, Mistral 7B, and Phi 3 mini-instruct models achieved better results than the baseline models BloombergGPT, AdaptLLM-7B and FinMA-7B, which are trained on financial data. Wang et al. \cite{b8} fine-tuned several open-source models such as Llama2 7B, Falcon 7B, BLOOM 7.1B, MPT 7B, ChatGLM2 6B and Qwen 7B with the FPB and FiQA datasets, where LLaMA2 7B and MPT 7B obtained the best results. These studies are summarized in Table \ref{tab:Literaturemodels}.

Since they rely on explicit instructions to capture sentiments, NLP models may encounter difficulties in extracting sentiments from some non-English languages such as Chinese. Stemming procedures to obtain the root, also called the stem, of words can be applied in English by removing suffixes. For example, one may extract the stem `look' by removing `ing' in `looking' or `ed' in `looked'. However, in Chinese, there is no concept of stem. Separating building blocks of Chinese characters would completely change their meaning. Without the need for explicit rules, LLMs can capture sentiments from Chinese LLMs by learning contextual representations via pre-training on massive corpora. Lan et al. (2023) \cite{b3} fine-tuned several LLMs, including Longformer, LLaMA, BLOOM and ChatGLM, on a Chinese sentiment dataset. Their results showed that Chinese LLaMA Plus outperformed other models in sentiment analysis, especially in a zero-shot learning scenario. However, no study considered addressing financial sentiment in both English and Chinese fine-tuning a single model.

\section{Methods}

This study aims to evaluate and compare the performance of four models, FinBERT, Llama3 8B, DeepSeek LLM 7B and Qwen3 8B, on financial data from multiple sources and language types. We adapt the training pipelines to align with the inherent constraints that vary among the investigated models. However, to ensure fair comparison, the training parameters remain the same, except for batch sizes due to memory limitations. We evaluated the models in a low-resource efficiency setting by varying training and test data ratios and asses how the LLMs perform with ZSL, FSL and SFT. The examples given in the prompt during the few-shot learning were randomly selected.

Without claiming to offer a comprehensive benchmark analysis of LLMs applied in sentiment analysis of financial data, our study attempts to offer a fair and transparent comparison of recent lightweight LLMs available publicly. The LLMs chosen have a similar number of parameters (7-8B) and were released between 2024 and 2025, their characteristics are summarized in Table \ref{tab:model_comparison}. 

\subsection{Data}
\begin{itemize}
    \item \textbf{FinancialPhraseBank (FBP):} A dataset of financial sentences labeled with sentiment categories, specifically designed for sentiment analysis in the financial domain \cite{b11}.
    \item \textbf{Financial Question Answering (FiQA):} A dataset containing financial question-answer pairs, labeled with sentiment annotations. It focuses on financial documents and provides essential data for training models to analyze sentiment in finance-related texts \cite{b12}.
    \item \textbf{Gold News Sentiment (GSD):} A sentiment-labeled dataset that focuses on news related to gold commodities \cite{b13}.
    \item \textbf{Twitter Sentiment (TSD):} A dataset consisting of tweets related to financial news, providing sentiment annotations. It allows analysis of sentiments expressed on social networks, offering a unique perspective compared to traditional news datasets \cite{b14}.
    \item \textbf{Chinese Finance Sentiment (CSD):} A dataset consisting of financial news articles labeled with sentiment annotations, specifically focused on Chinese financial news, which allows sentiment classification tasks based on text in Chinese language \cite{b15}.
\end{itemize}

\begin{figure}[t]
       \centering
          \includegraphics[width=1\textwidth]{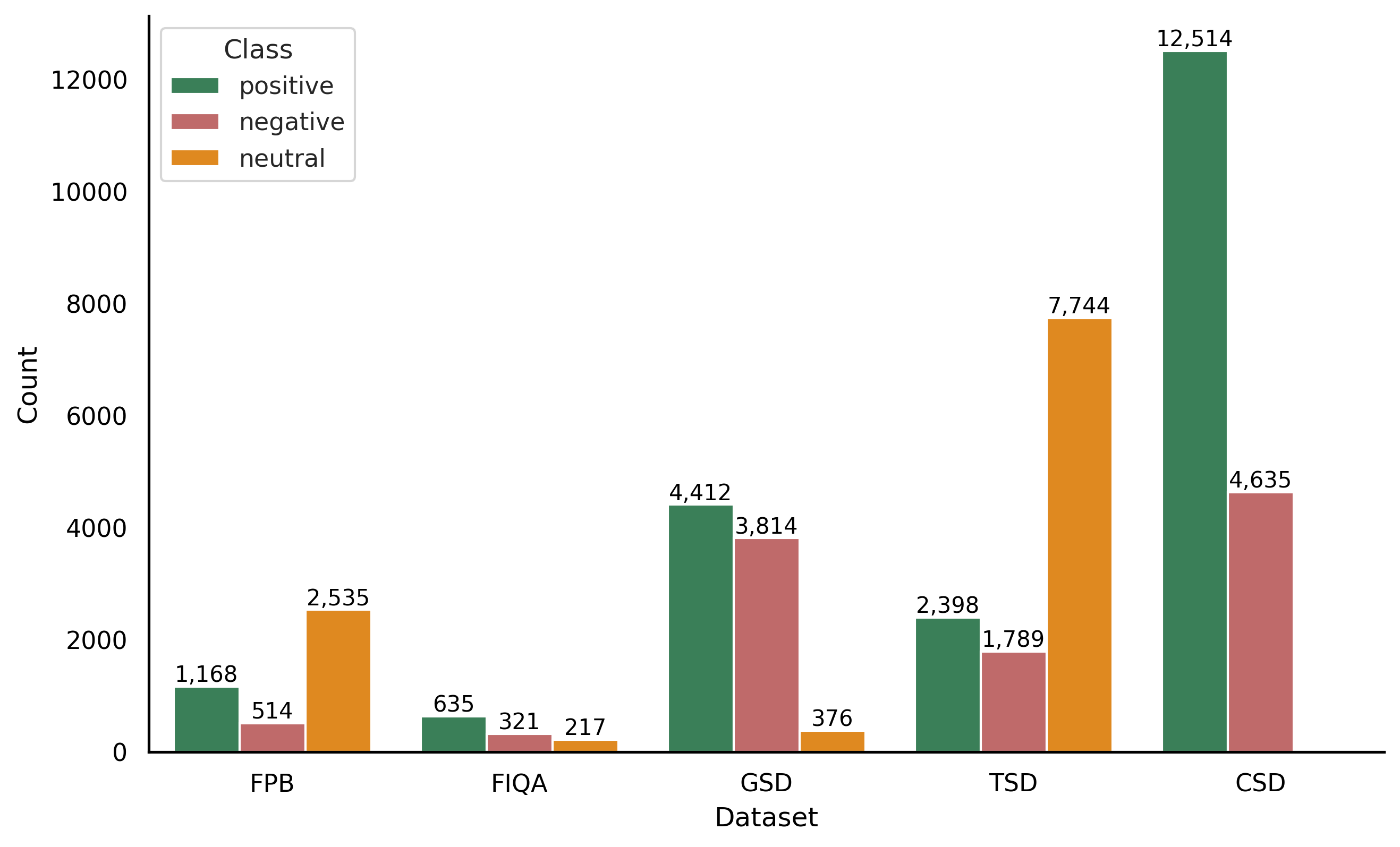}
          \caption{Class distribution per dataset. FinancialPhraseBank (FPB), Financial Question Answering (FiQA), Gold News Sentiment (GSD), Twitter Sentiment (TSD) and Chinese Sentiment (CSD).}
          \label{fig:classes}
\end{figure}

FinancialPhraseBank, FiQA, Gold News Sentiment, Twitter Sentiment are available on Huggingface, and Chinese Finance Sentiment is available in \cite{b15}. All labels in the datasets are unbalanced (Fig. \ref{fig:classes}), which means that the proportion of observations between the sentiment classes (positive, negative, and neutral) differs substantially. Unbalanced data are common in the financial sector, such as bankruptcy prediction, credit card fraud, and credit approval, and therefore require care in analyzing them \cite{b22}.
The choice of predictive scores is particularly important in this context. Accuracy, which refers to the proportion of correct predictions made by a model out of all predictions, may be misleading. A trivial classifier that would always predict the majority class consisting of, for example, 90\% of the labels would reach the accuracy of 90\%. To address this issue, we will complement the accuracy with the macro F1 score, which accounts for the class imbalance in assessing the predictive performance of the models. 

\subsection{Workflow}
We designed a pipeline to carry out the sentiment analysis on financial multi-domain datasets, such as tweets, headlines, and microblogs, which vary by their language style and format. The pipeline allows models to train on multiple datasets simultaneously and fine-tune the five investigated financial sentiment datasets, combining English and Chinese data. Based on an empirical analysis on similar datasets (see details in Appendix A.3), we opted for a balanced learning across datasets and domains. This procedure automatically sub-samples or up-samples when needed to maintain equal representation and prevents bias towards more frequent domains. Here, we iteratively varied the proportions of the training data, from 5\% to a fully 100\% fine-tuned model.  

The training process follows three phases. During the initialization phase (20\% of the steps), the model prioritizes underrepresented domains (e.g., FiQA) with learning rate warm-up, which is a technique that gradually increases the learning rate at the beginning of training to stabilize and improve model convergence. In the balanced phase (60\% of the steps), the training data is split into small subsets of data (called batches) to ensure that each type of data (or domain) is equally represented by applying weights. These weights control the sampling probability of each dataset, so that smaller domains with fewer examples are oversampled (their samples appear more frequently), while larger domains are undersampled (a smaller fraction of their samples is drawn per epoch). This weighted sampling guaranties that on average each domain contributes a similar number of updates to the model during training and prevents the learning process from being dominated by one type over another. The learning rate follows the decay of the cosine learning rate. This is a scheduling technique where the learning rate decreases over time following a cosine curve. In other words, it starts higher and gradually slows down. This helps the model learn quickly at first, and after the model has learned a lot, it slows down, allowing it to make smaller adjustments. The finalization phase (20\% of the steps) consists of a reduction in the learning rate per layer and the application of early stopping per domain.

To fine-tune the LLMs, we applied Low-Rank Adaptation (LoRA \cite{b28}) and quantization. The LoRA method is facilitated by the Parameter Efficient Fine-Tuning (PEFT) module, which enables fine-tuning large language models by updating only a small subset of parameters, such as adapters or LoRA layers, making training faster, cheaper, and more memory efficient. Quantification is done via the BitsAndBytes \cite{b29} library, which enables efficient model training and inference by providing low-bit (e.g., 4-bit, 8-bit) quantization and optimized CUDA kernels to reduce memory usage and accelerate performance. The chosen quantization setup was 4-bit for weight storage and bfloat16 for the computation dtype for fine-tuning and inference. In this way, we can achieve a higher accuracy using bfloat16 during the operations calculation but with a reduced GPU memory usage from the model loading. A single NVIDIA A100 40GB GPU is used for fine-tuning and inference.

\section{Results and discussion}
\subsection{Performance assessment on financial sentiment datasets}
We present a comprehensive evaluation of the performance of FinBERT, DeepSeek LLM 7B, Llama3 8B Instruct, and Qwen3 8B in predicting sentiments based on four financial datasets in English. Table \ref{tab:llm-shots-comparison} shows the predictive performance (accuracy and macro F1 score) of a first evaluation, which includes only the LLM in 0 shot, 3 shot and 5-shot settings. Both results of 3-shot and 5-shot settings are averaged because of the randomness when choosing examples provided in the prompt, which come from the test data. 

\begin{table}[ht]
\centering
\scriptsize
\begin{tabular}{lllccl}
\toprule
\textbf{Model} & \textbf{Shots} & \textbf{Dataset} & \textbf{Acc.} & \textbf{Macro F1} \\
\midrule
\multirow{15}{*}{DeepSeek LLM 7B} & 
    \multirow{5}{*}{0-shot}
        & FiQA & 0.64 & 0.42 \\
        & & FPB  & 0.34 & 0.38 \\
        & & GSD  & 0.69 & 0.55 \\
        & & TSD  & 0.27 & 0.30 \\
        & & CSD  & 0.86 & 0.58 \\
\cmidrule(lr){2-5}
 & \multirow{5}{*}{3-shot} & FiQA & $0.60\pm0.02$ & $0.41\pm0.03$ \\
 & & FPB  & $0.38\pm0.01$ & $0.39\pm0.01$ \\
& & GSD  & $0.61\pm0.01$ & $0.41\pm0.01$ \\
 & & TSD  & $0.33\pm0.02$ & $0.30\pm0.01$ \\
  & & CSD  & $0.61\pm0.02$ & $0.45\pm0.09$ \\
\cmidrule(lr){2-5}
 & \multirow{5}{*}{5-shot} & FiQA & $0.60\pm0.03$ & $0.47\pm0.05$ \\
 & & FPB  & $0.40\pm0.01$ & $0.47\pm0.05$ \\
 & & GSD  & $0.58\pm0.01$ & $0.39\pm0.01$ \\
 & & TSD  & $0.38\pm0.00$ & $0.36\pm0.02$ \\
 & & CSD  & $0.54\pm0.01$ & $0.47\pm0.02$ \\
\midrule
\multirow{15}{*}{LLaMA3 8B Instruct} 
 & \multirow{5}{*}{0-shot} & FiQA & 0.53 & 0.53 \\
 &  & FPB  & 0.65 & 0.39 \\
 &  & GSD  & 0.17 & 0.19 \\
 &  & TSD  & 0.75 & 0.64 \\
 &  & CSD  & 0.65 & 0.42 \\
\cmidrule(lr){2-5}
 & \multirow{5}{*}{3-shot} & FiQA & $0.49\pm0.02$ & $0.48\pm0.01$ \\
 & & FPB  & $0.76\pm0.01$ & $0.71\pm0.01$ \\
 & & GSD  & $0.33\pm0.01$ & $0.33\pm0.00$ \\
 & & TSD  & $0.66\pm0.00$ & $0.63\pm0.00$ \\
 & & CSD  & $0.82\pm0.01$ & $0.81\pm0.01$ \\
 \cmidrule(lr){2-5}
 & \multirow{5}{*}{5-shot} & FiQA & $0.59\pm0.02$ & $0.55\pm0.03$ \\
 & & FPB  & $0.62\pm0.01$ & $0.61\pm0.01$ \\
 & & GSD  & $0.42\pm0.00$ & $0.40\pm0.00$ \\
 & & TSD  & $0.59\pm0.00$ & $0.57\pm0.01$ \\
 & & CSD  & $0.84\pm0.01$ & $0.84\pm0.01$ \\
\midrule
\multirow{15}{*}{Qwen3 8B} 
 & \multirow{5}{*}{0-shot} & FiQA & 0.29 & 0.31 \\
 & & FPB  & 0.82 & 0.78 \\
 & & GSD  & 0.56 & 0.51 \\
 & & TSD  & 0.79 & 0.73 \\
 & & CSD  & 0.91 & 0.61 \\
\cmidrule(lr){2-5}
 & \multirow{5}{*}{3-shot} & FiQA & $0.43\pm0.01$ & $0.45\pm0.01$ \\
 & & FPB  & $0.84\pm0.00$ & $0.80\pm0.00$ \\
 & & GSD  & $0.69\pm0.00$ & $0.61\pm0.00$ \\
 & & TSD  & $0.81\pm0.00$ & $0.74\pm0.01$ \\
 & & CSD  & $0.92\pm0.00$ & $0.61\pm0.00$ \\
\cmidrule(lr){2-5}
 & \multirow{5}{*}{5-shot} & FiQA & $0.55\pm0.02$ & $0.51\pm0.01$ \\
 & & FPB  & $0.66\pm0.02$ & $0.65\pm0.02$ \\
 & & GSD  & $0.75\pm0.01$ & $0.65\pm0.01$ \\
 & & TSD  & $0.75\pm0.01$ & $0.70\pm0.01$ \\
 & & CSD  & $0.94\pm0.00$ & $0.63\pm0.00$ \\


\bottomrule
\end{tabular}
\caption{Performance comparison of DeepSeek LLM 7B, LLaMA3 8B Instruct, and Qwen3 8B across datasets under 0-shot, 3-shot, and 5-shot settings. Metrics: Accuracy (Acc.) and macro F1.}
\label{tab:llm-shots-comparison}
\end{table}

The results show that Qwen and Llama models benefit most from few-shot learning, with performance improving markedly from 0-shot to 3-shot settings for the datasets FPB, GSD and CSD. Changes from 3-shot to 5-shot learning do not appear to substantially change the model performance. Deepseek shows worse results in 3 and 5-shot settings across all datasets except for TSD, where they increase slightly. The small standard deviations of accuracy and macro f1 suggest that using different examples inside the prompt does not have such a big impact on the model prediction performance. Overall, Qwen achieves the best results on 0-shot with an average accuracy and macro f1 of 0.67 and 0.59 respectively, while Deepseek achieves 0.67 and 0.55 and Llama 0.55 and 0.43. For 3-shot, Qwen gets the best results again, 0.74 accuracy and 0.64 macro f1 score, and for 5-shot, Qwen is still the best performer with 0.73 and 0.63 average accuracy and macro f1 scores.

Following this initial assessment, we fine-tuned the models on varying data proportions (5, 10, 20, 40, 75 and 100\%). FinBERT's prior pre-training with FiQA and FPB datasets forbids its re-evaluation on those to avoid data leakage. Table \ref{tab:sft-model-comparison} shows the predictive performance (accuracy and macro F1 score) of this second evaluation, which includes all models with various proportions of training data.

\begin{table}[ht]
\centering
\scriptsize
\begin{tabular}{llcccccc}
\toprule
\textbf{Dataset} & \textbf{Metric} & \textbf{5\%} & \textbf{10\%} & \textbf{20\%} & \textbf{40\%} & \textbf{75\%} & \textbf{100\%} \\
\midrule
\multicolumn{8}{c}{\textbf{FinBERT}} \\
\midrule
\multirow{2}{*}{GSD} & Acc. & 0.82 & 0.79 & 0.76 & 0.70 & 0.58 & 0.56 \\
 & Macro F1 & 0.65 & 0.63 & 0.58 & 0.58 & 0.45 & 0.45 \\
\multirow{2}{*}{TSD} & Acc. & 0.73 & 0.69 & 0.52 & 0.71 & 0.64 & 0.68 \\
 & Macro F1 & 0.67 & 0.64 & 0.52 & 0.61 & 0.57 & 0.55 \\
\midrule
\multicolumn{8}{c}{\textbf{DeepSeek LLM 7B}} \\
\midrule
\multirow{2}{*}{GSD} & Acc. & 0.58 & 0.88 & 0.91 & 0.93 & 0.93 & 0.94 \\
 & Macro F1 & 0.50 & 0.76 & 0.79 & 0.84 & 0.87 & 0.88 \\
\multirow{2}{*}{TSD} & Acc. & 0.57 & 0.72 & 0.82 & 0.85 & 0.87 & 0.88 \\
 & Macro F1 & 0.51 & 0.68 & 0.78 & 0.81 & 0.84 & 0.85 \\
\multirow{2}{*}{CSD} & Acc. & 0.79 & 0.95 & 0.96 & 0.96 & 0.97 & \textbf{0.97} \\
 & Macro F1 & 0.79 & 0.95 & 0.96 & 0.96 & 0.96 & \textbf{0.97} \\
\multirow{2}{*}{FIQA}   & Acc. & 0.57 & 0.71 & 0.77 & 0.73 & 0.79 & 0.84 \\
 & Macro F1 & 0.41 & 0.60 & 0.67 & 0.62 & 0.70 & 0.77 \\
\multirow{2}{*}{FPB} & Acc. & 0.42 & 0.68 & 0.85 & 0.87 & 0.91 & 0.92 \\
 & Macro F1 & 0.46 & 0.70 & 0.82 & 0.85 & 0.90 & 0.91 \\
\midrule
\multicolumn{8}{c}{\textbf{Qwen3 8B}} \\
\midrule
\multirow{2}{*}{GSD} & Acc. & 0.88 & 0.90 & 0.92 & 0.93 & 0.94 & \textbf{0.95} \\
 & Macro F1 & 0.78 & 0.82 & 0.86 & 0.87 & 0.87 & \textbf{0.88} \\
\multirow{2}{*}{TSD} & Acc. & 0.81 & 0.85 & 0.86 & 0.87 & 0.89 & 0.89 \\
 & Macro F1 & 0.78 & 0.81 & 0.82 & 0.85 & 0.86 & 0.86 \\
\multirow{2}{*}{CSD} & Acc. & 0.95 & 0.96 & 0.96 & 0.97 & 0.97 & \textbf{0.97} \\
 & Macro F1 & 0.95 & 0.96 & 0.96 & 0.97 & 0.97 & \textbf{0.97} \\
\multirow{2}{*}{FIQA} & Acc. & 0.81 & 0.83 & 0.80 & 0.82 & 0.83 & \textbf{0.84} \\
 & Macro F1 & 0.73 & 0.72 & 0.72 & 0.74 & 0.74 & \textbf{0.77} \\
\multirow{2}{*}{FPB} & Acc. & 0.85 & 0.88 & 0.91 & 0.91 & 0.92 & \textbf{0.93} \\
 & Macro F1 & 0.82 & 0.87 & 0.90 & 0.90 & 0.91 & \textbf{0.92} \\
\midrule
\multicolumn{8}{c}{\textbf{Llama3 8B}} \\
\midrule
\multirow{2}{*}{GSD} & Acc. & 0.89 & 0.92 & 0.93 & 0.94 & 0.94 & \textbf{0.95} \\
 & Macro F1 & 0.79 & 0.82 & 0.86 & 0.87 & 0.87 & \textbf{0.88} \\
\multirow{2}{*}{TSD} & Acc. & 0.82 & 0.85 & 0.85 & 0.88 & 0.89 & \textbf{0.89} \\
 & Macro F1 & 0.78 & 0.81 & 0.82 & 0.85 & 0.86 & \textbf{0.87} \\
\multirow{2}{*}{CSD} & Acc. & 0.95 & 0.96 & 0.96 & 0.96 & 0.96 & 0.96 \\
 & Macro F1 & 0.95 & 0.96 & 0.96 & 0.96 & 0.96 & 0.96 \\
\multirow{2}{*}{FIQA} & Acc. & 0.81 & 0.83 & 0.80 & 0.82 & 0.83 & \textbf{0.84} \\
 & Macro F1 & 0.73 & 0.71 & 0.72 & 0.74 & 0.74 & \textbf{0.77} \\
\multirow{2}{*}{FPB} & Acc. & 0.85 & 0.88 & 0.91 & 0.91 & 0.92 & \textbf{0.93} \\
 & Macro F1 & 0.82 & 0.87 & 0.90 & 0.90 & 0.91 & \textbf{0.92} \\
\bottomrule
\end{tabular}
\caption{\centering{Comparison of FinBERT, DeepSeek, Qwen3, and Llama3 across different training data proportions and datasets (FiQA, FPB, GSD, TSD, CSD). Metrics reported: Accuracy and macro F1 score. Best results in \textbf{bold}.}}
\label{tab:sft-model-comparison}
\end{table}

FinBERT consistently underperformed, with accuracy and macro F1 scores decreasing for GSD (up to a -30\% accuracy and macro F1 score) and TSD (-18\% macro F1 score) as it is fine-tuned. This is most likely the result of overfitting during training or catastrophic interference. In contrast, LLMs show great performance with a bare minimum of data used to fine-tune (5\%), except for DeepSeek evaluated in FPB and TSD datasets; however, its performance jumps quickly with training data of 10\%. Qwen exhibited robust performance across all datasets, maintaining high accuracy and macro F1 scores even at low data proportions, suggesting efficient learning and generalization. In the fully fine-tuned stage, Llama slightly surpassed the Qwen and Deepseek models, achieving the highest overall performance across most datasets and training sizes. These findings suggest that, while FinBERT is limited in its generalization beyond its pretraining domain, LLMs offer superior scalability and sentiment classification capabilities, making them ideal for both low- and high-data resource applications.

\begin{figure}[t]
       \centering
          \includegraphics[width=1\textwidth]{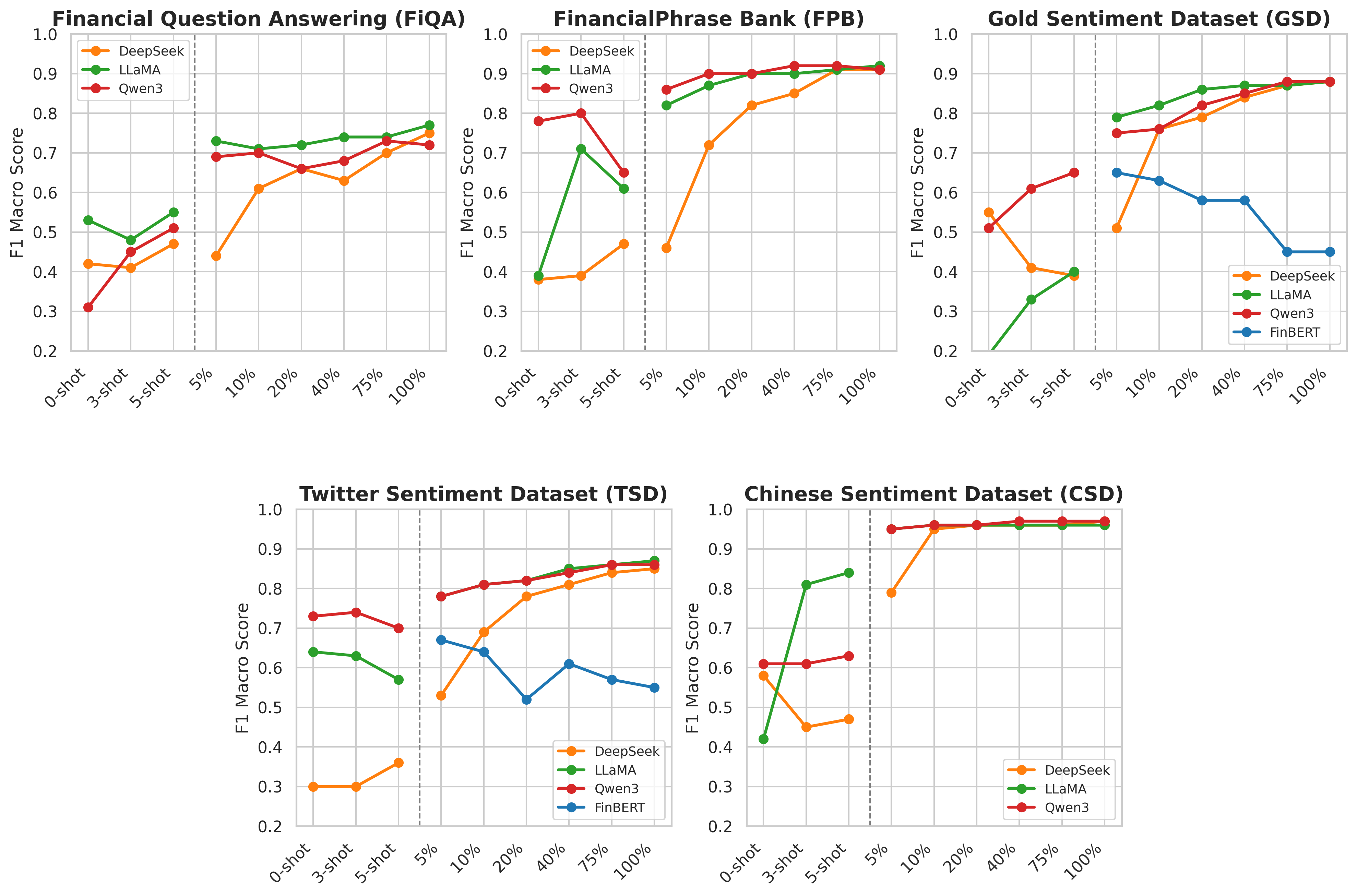}
          \caption{Fine-tuned and 0, 3, 5-shot learning results for all models. X-axis is shots used or training data proportion used for finetuning. Y-axis is F1 Macro score. In orange, green, red and blue are the results of DeepSeek, Llama, Qwen and FinBERT models respectively.}
\label{fig:final-results}
\end{figure}

The three LLMs show the worst performance when tested on the FiQA dataset, most likely due to the small size of this dataset. It has only around 1200 sentences, in contrast with the total size of the dataset used to fine-tune these models formed by the 4 datasets together, adding to a total of almost 26000 sentences. This can cause interference in the models' weights, which happens when models forget previously learned information while training with a new one. Furthermore, they all achieve up to 0.97 F1 Score when tested on the Chinese Sentiment Dataset. These results are higher likely because this dataset only contains two classes, positive and negative, instead of three as the rest. The results with balanced training (ZSL and FSL) and fine-tuned evaluations are shown in Fig. \ref{fig:final-results}.

We compare the results of the best prediction model (and an alternative specification that used only 10\% of the data for training) with benchmark models from previous studies on the same datasets (Table \ref{tab:literature-model-performance}). The best model shows more consistent results across different financial sentiment datasets in English language, even having been fine-tuned in a large proportion with Chinese language data. In addition, Llama3 model fine-tuned with only 10\% of the data still performed competitively or even outperformed the models in previous works.
\begin{table}[t]
\centering
\scriptsize
\begin{tabular}{c l c c c c c c}
\toprule
\textbf{No.} & \textbf{Model} & \textbf{Metric} & \textbf{FPB} & \textbf{TSD (test)} & \textbf{FiQA} & \textbf{GSD} & \textbf{Average} \\
\midrule
\multirow{2}{*}{1} & \multirow{2}{6cm}{SFT LLaMA-7B \cite{b4}} & Acc & 0.76 & 0.86 & - & - & 0.81 \\
                   & & F1 * & 0.74 & 0.81 & - & - & 0.78 \\
\midrule
\multirow{2}{*}{2} & \multirow{2}{6cm}{SFT Flan-T5 XL \cite{b5}} & Acc & 0.81 & \textbf{0.90} & - & - & 0.86 \\
                   & & Macro F1 & 0.78 & \textbf{0.88} & - & - & 0.83 \\
\midrule
\multirow{2}{*}{3} & \multirow{2}{6cm}{Instruct-FinGPT-7B (SFT LLaMA-7B) \cite{b7}} & Acc & 0.76 & 0.88 & - & - & 0.82 \\
                   & & F1 * & 0.74 & 0.84 & - & - & 0.79 \\
\midrule
\multirow{2}{*}{4} & \multirow{2}{6cm}{SFT MPT \cite{b8}} & Acc & - & - & - & - & - \\
                   & & F1 * & 0.87 & - & \textbf{0.86} & - & 0.87 \\
\midrule
\multirow{2}{*}{5} & \multirow{2}{6cm}{SFT Mistral-7B (FPB) \& LLaMA3-8B (FiQA) \cite{b9}} & Acc & 0.87 & - & \textbf{0.87} & - & 0.87 \\
                   & & Weighted F1 & 0.87 & - & \textbf{0.85} & - & 0.86 \\
\midrule
\multirow{2}{*}{6} & \multirow{2}{6cm}{Random Forest Classifier \cite{b10}} & Acc & - & - & - & - & - \\
                   & & F1 * & - & - & - & 0.73 & 0.73 \\
\midrule
\multirow{4}{*}{\rotatebox[origin=c]{90}{Our models}} 
& \multirow{2}{6cm}{SFT Llama3 8B Instruct} & Acc & \textbf{0.93} & 0.89 & 0.84 & \textbf{0.95} & \textbf{0.90} \\
& & Macro F1 & 0.92 & 0.87 & 0.84** & \textbf{0.88} & \textbf{0.88} \\
\cmidrule(lr){2-8}
& \multirow{2}{6cm}{10\% SFT Llama3 8B***} & Acc & 0.88 & 0.85 & 0.83 & 0.92 & 0.87 \\
& & Macro F1 & 0.87 & 0.81 & 0.83** & 0.82 & 0.83 \\
\bottomrule
\end{tabular}
\caption{\centering Performance comparison between lightweight LLMs and the literature in classifying sentiment. SFT means supervised fine-tuned. *F1 score was reported without specifying whether it was Macro, Micro or Averaged, so we considered it macro for comparison. **This represented weighted F1 to align with the paper \#5. ***40\% SFT means the model was fine-tuned on only 40\% of the training data. Bold values highlight the best results. The results of both our fully fine-tuned and 10\% fine-tuned offer more solid results than previous works and show how minimal training datasets can be enough to achieve near peak performance.}
\label{tab:literature-model-performance}
\end{table}

\section{Conclusion}
We perform a comparative analysis between FinBERT, a domain-specific natural language processing model for financial sentiment analysis, and parameter efficient large language models, with the objective of predicting sentiment in financial text corpora. We assess the low-resource capabilities of LLMs by fine-tuning with different proportions of the training data as well as evaluating the base models in 0, 3 and 5 shot learning in English and Chinese languages. We find that large language models (LLMs) can surpass the performance of FinBERT while relying on only a minimal proportion of training data, or even in the complete absence of task-specific training data under a zero-shot evaluation setting.

Secondly, we design a domain-balanced training pipeline to improve the fine-tuning process by considering the training data as a corpus of datasets from various sources and sizes. The results suggest that fine-tuning the models in such a way may improve predictive performance compared to a sequential training strategy.

Future work may explore incorporating additional financial datasets, fine-tuning of LLMs on multilingual sentiment datasets, extending to sentiment tasks outside of finance, or integrating RAG for more contextual predictions. We hope that the tools, methods, and findings presented here will support researchers and practitioners in building robust sentiment analysis models tailored to the financial domain, particularly in data-scarce and low-resource environments.

\clearpage
\appendix
\section*{Appendix}
\addcontentsline{toc}{section}{Appendix}
\section{Previous studies using these datasets}
Here we list previous studies using financial textual data and indicate how the data was used (training and/or test) \ref{tab:Literaturemodels}.
\begin{table}[h]
\centering
\scriptsize
\begin{tabular}{c c p{4.5cm} p{1.5cm} p{2.5cm} p{4.5cm}}
\toprule
\textbf{No.} & \textbf{Year} & \textbf{Title} & \textbf{Authors} & \textbf{Data Used} & \textbf{Models} \\
\midrule
1 & 2023 & Enhancing Financial Sentiment Analysis via Retrieval Augmented Large Language Models \cite{b4} & Zhang et al. & TSD$^{\text{tr, te}}$, FiQA$^{\text{tr}}$, FPB$^{\text{te}}$ & LLaMA-7B (theirs); BloombergGPT, ChatGPT, LLaMA-7B, ChatGLM2-6B, FinBERT \\
\midrule
2 & 2023 & A Comparative Analysis of Fine-Tuned LLMs and Few-Shot Learning of LLMs for Financial Sentiment Analysis \cite{b5} & Fatemi et al. & TSD$^{\text{tr, te}}$, FPB$^{\text{te}}$ & Flan-T5-Base/Large/XL (248M, 785M, 2.85B) (theirs); FinBERT, Instruct-FinGPT-7B, ChatGPT \\
\midrule
3 & 2023 & Instruct-FinGPT: Financial Sentiment Analysis by Instruction Tuning of General-Purpose Large Language Models \cite{b7} & Zhang et al. & TSD$^{\text{tr, te}}$, FiQA$^{\text{tr}}$, Numerical sensitivity$^{\text{te}}$, Contextual understanding$^{\text{te}}$, FPB$^{\text{te}}$ & LLaMA-7B (theirs); LLaMA-7B, FinBERT, ChatGPT \\
\midrule
4 & 2023 & FinGPT: Instruction Tuning Benchmark for Open-Source Large Language Models in Financial Datasets \cite{b8} & Wang et al. & FPB$^{\text{tr, te}}$, FiQA$^{\text{tr, te}}$, TFNS$^{\text{tr, te}}$, NWGI$^{\text{tr, te}}$ & LLaMA2-7B, Falcon-7B, BLOOM-7.1B, MPT-7B, ChatGLM2-6B, Qwen-7B (theirs and benchmarks) \\
\midrule
5 & 2024 & A Comparative Analysis of Instruction Fine-Tuning LLMs for Financial Text Classification \cite{b9} & Fatemi et al. & FPB$^{\text{tr, te}}$, FiQA$^{\text{tr, te}}$ & LLaMA3-8B, Mistral-7B, Phi-3-mini-Instruct (theirs); BloombergGPT, AdaptLLM-7B, FinMA-7B, GPT-4 \\
\midrule
6 & 2024 & Analyzing Market Sentiment for Gold Commodity
News Through Natural Language Processing
Techniques \cite{b10} & Bora et al. & GSD$^{\text{tr, te}}$ & Random Forest Classifier \\
\bottomrule
\end{tabular}
\caption{\centering Overview of prior works using financial sentiment analysis datasets. In “Data Used”, \textit{tr} and \textit{te} indicate use for training and testing, respectively.}
\label{tab:Literaturemodels}
\end{table}

\section{Investigated LLMs}

FinBERT is a BERT-based model that has already been fine-tuned with several financial datasets, including the Financial PhraseBank dataset, the FiQA dataset, Reuters-21578, SEC Filings (10-K/10-Q Reports) and other financial news articles and financial-related Twitter data. It directly outputs sentiment and confidence scores and is specifically optimized for sentiment analysis in financial domains.

DeepSeek-LLM-7B is a 7 billion-parameter open-weight large language model developed by DeepSeek AI, optimized for efficiency and high performance in NLP tasks such as sentiment analysis. It incorporates grouped-query attention (GQA) and a mixture-of-experts (MoE) architecture to enhance inference speed while maintaining strong contextual understanding.

LLaMA-3-8B-Instruct, released by Meta, is an 8 billion-parameter instruction-tuned variant of the LLaMA-3 series, designed for conversational and task-oriented applications through supervised fine-tuning (SFT) and reinforcement learning from human feedback (RLHF). 

Qwen3-8B is an 8 billion-parameter decoder-based LLM developed by Alibaba’s Qwen team, offering enhanced reasoning and multilingual support, particularly for English and Chinese. With an extended 32k token context window, it excels in long-context sentiment analysis tasks such as product reviews or social media threads. The open weight release and fine-tuning flexibility of the model make it suitable for comparing its performance with models of similar sizes.

\begin{table}[t]
\centering
\scriptsize
\begin{tabular}{lcccl}
\toprule
\textbf{Model} & \textbf{Year} & \textbf{Developer} & \textbf{Parameters} & \textbf{Languages Supported} \\
\midrule
DeepSeek-LLM-7B & 2023 & DeepSeek AI & 7B & English, Chinese, Multilingual \\
LLaMA-3-8B-Instruct & 2024 & Meta (Facebook) & 8B & Primarily English \\
Qwen3-8B & 2025 & Alibaba & 8B & English, Chinese + Multilingual \\
\bottomrule
\end{tabular}
\caption{Comparison of Open-Weight LLMs used for Sentiment Analysis}
\label{tab:model_comparison}
\end{table}

\section{Fine-tuning strategy}
\begin{figure}[t]
       \centering
          \includegraphics[width=1\textwidth]{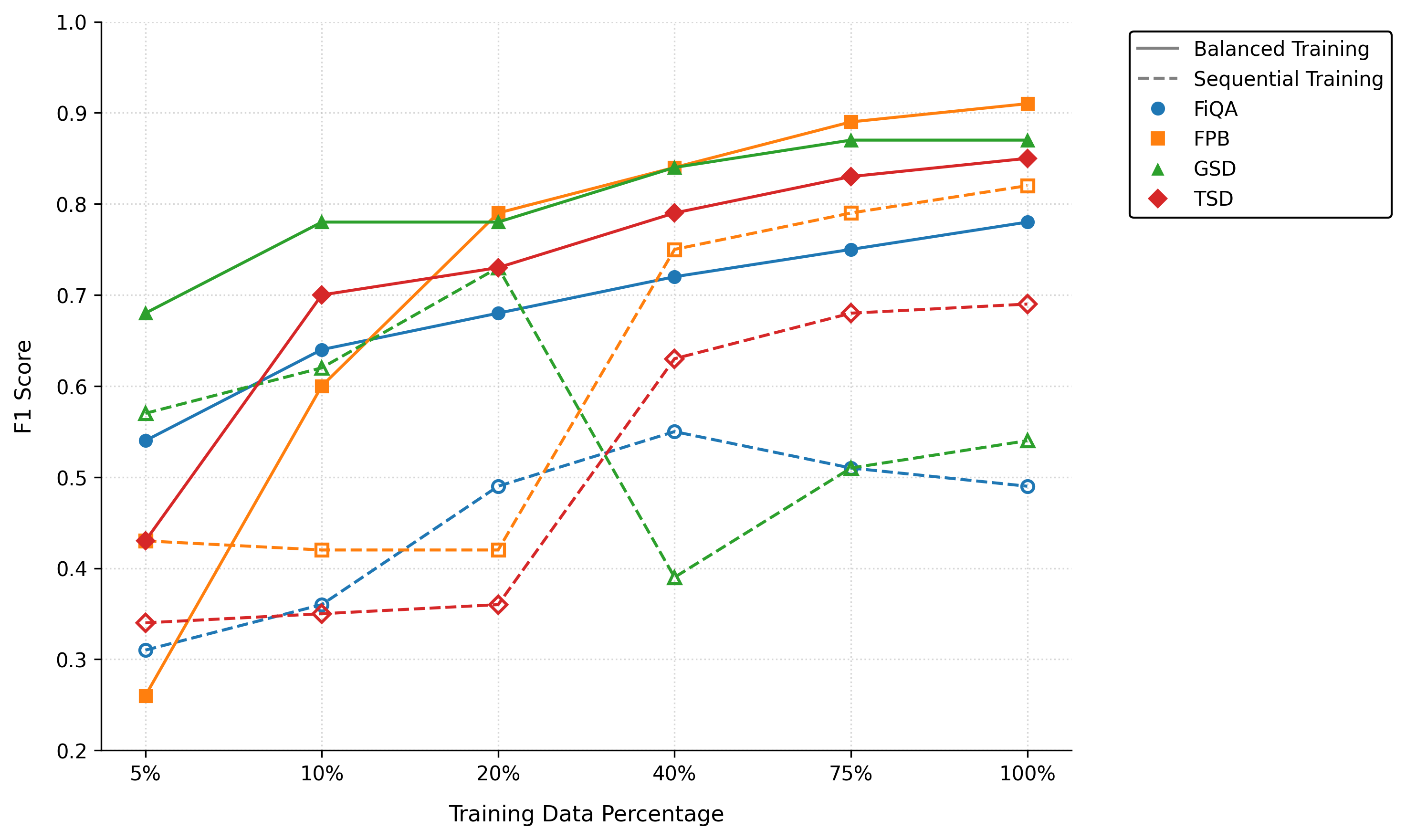}
          \caption{\textbf{Comparison of training strategies applied to predict sentiment} Illustration of the predictive performance of Deepseek LLM 7B using sequential (dashed lines) and balanced (solid lines) fine-tuning methods. The x-axis represents the proportion of data used for training (from 5\% to 100\%) and the y-axis shows the macro F1 score.}
\label{sequential_balanced_fig}
\end{figure}

We compare the performance in predicting sentiment based on datasets FinancialPhraseBank (FPB), Financial Question Answering (FiQA), Gold News Sentiment (GSD) and Twitter Sentiment (TSD) and Deepseek LLM 7B with balanced (solid lines) and sequential (dashed lines) training. The results show a systematic improvement of the performance when balanced training is applied (Fig. \ref{sequential_balanced_fig}), which supports our methodological procedure choice.  

\FloatBarrier





\end{document}